%% file: main.tex
\documentclass[]{OptAI}

\usepackage{amsmath} 
\usepackage{natbib}
\usepackage{graphicx}
\usepackage{subcaption} 

\usepackage[toc,page,header]{appendix}
\usepackage[utf8]{inputenc} 
\usepackage[T1]{fontenc}    

\usepackage{url}            
\usepackage{booktabs}       
\usepackage{pifont}        
\newcommand{\xmark}{\ding{55}} 
\newcommand{\cmark}{\ding{51}} 
\usepackage{amsfonts}       
\usepackage{nicefrac}       
\usepackage{microtype}      
\usepackage{wrapfig}

\usepackage{amssymb}  
\usepackage{fontawesome}  
\usepackage{url}  

\usepackage{titletoc}

\usepackage{tikz}  
\usepackage{comment}  
\usepackage{tabularx}  
\usepackage{booktabs}  

\usepackage{minitoc}

\usepackage{booktabs}
\usepackage{array}
\usepackage{etoolbox}

\definecolor{lightblue}{RGB}{200, 230, 255}  
\definecolor{headerblue}{RGB}{150, 200, 255} 

\usepackage{pgfplots}
\usepackage[utf8]{inputenc} 
\usepackage[T1]{fontenc}    

\usepackage{url}            
\usepackage{booktabs}       
\usepackage{amsfonts}       
\usepackage{nicefrac}       
\usepackage{microtype}      
\usepackage{xcolor}         
\usepackage{graphicx}
\usepackage{float}
\usepackage{comment}
\usepackage{multirow} 
\usepackage{amsmath} 
\usepackage{makecell} 
\usepackage{siunitx}  
\usepackage{tikz}
\usepackage{pgf-pie} 
\usepackage{subcaption}
\usepackage{wrapfig}
\usepackage[export]{adjustbox}

\usepackage{ragged2e}      
\usepackage{tabularx}       
\usepackage{array}          
\usepackage{caption}        
\usepackage{enumitem}
\usepackage{pifont}
\usepackage[hang,flushmargin]{footmisc} 

\usepackage{tcolorbox}

\usepackage{tcolorbox}    
\tcbuselibrary{breakable}  
\tcbuselibrary{skins}      

\usepackage{tabularx}
\usepackage{listings}

\usepackage{times}
\usepackage{latexsym}

\usepackage[T1]{fontenc}

\usepackage[utf8]{inputenc}

\usepackage{microtype}

\usepackage{inconsolata}

\usepackage{graphicx}

\usepackage{amsmath}
\usepackage{amsfonts}

\usepackage[most]{tcolorbox}
\usepackage{listings}
\usepackage{ragged2e}
\usepackage{enumitem}
\usepackage{booktabs}
\usepackage[most]{tcolorbox}
\usepackage{listings}
\usepackage{ragged2e}
\tcbuselibrary{breakable,listings}

\usepackage{algorithm}
\usepackage{algorithmic}
\setboolean{ALC@noend}{true}
\usepackage{cuted} 

\definecolor{codegreen}{rgb}{0,0.6,0}
\definecolor{codegray}{rgb}{0.5,0.5,0.5}
\definecolor{codepink}{RGB}{252, 142, 172}
\definecolor{codepurple}{rgb}{0.58,0,0.82}
\definecolor{backcolour}{RGB}{245,245,245}
\lstdefinestyle{scaler}{
    backgroundcolor=\color{backcolour},   
    commentstyle=\color{magenta},
    keywordstyle=\color{blue},
    numberstyle=\tiny\color{codegray},
    stringstyle=\color{codepurple},
    basicstyle=\fontfamily{\ttdefault}\footnotesize,
    breakatwhitespace=false,        
    breaklines=true,                
    keepspaces=true,    
    frame=single,
    numbersep=5pt,                  
    showspaces=false,              
    showstringspaces=false,
    showtabs=false,               
    tabsize=2,
    classoffset=1, 
    keywordstyle=\color{violet},
    classoffset=0,
}
\newtcblisting{PromptBox}[1]{%
enhanced,
  breakable,
  listing only,
  colback=white,
  colframe=black!60,
  boxrule=0.8pt,
  arc=2pt,
  left=6pt,right=6pt,top=6pt,bottom=6pt,
  colbacktitle=black!60,
  coltitle=white,
  fonttitle=\bfseries,
  title={#1},
  before upper=\RaggedRight,
  segmentation style={draw=none},
  listing engine=listings,
  listing options={style=prompt}
}
\usepackage{amsmath}
\usepackage{array}
\usepackage{longtable} 

\usepackage{amsmath}
\usepackage{amssymb}
\usepackage{bm}
\usepackage{algorithm}
\usepackage{algorithmic}
\usepackage{longtable}
\usepackage[most]{tcolorbox}

\usepackage{enumitem}
\usepackage{multirow}
\usepackage{booktabs}

\newtcolorbox[
  auto counter
]{mybox}[1][]{%
  colback=blue!5!white,
  colframe=blue!75!black,
  arc=4pt,
  boxrule=0.6pt,
  left=6pt, right=6pt, top=4pt, bottom=4pt,
  title={},
  #1
}

\title{Opt.Gear Technical Report}

\author{
 \textbf{Opt.Gear Team$^\dagger$}
}

\input{commands.tex}

\abstract{
\begin{abstract}

We introduce Opt.Gear, a foundation model designed for efficient on-device deployment, real-tim inference, and strong task capability. It includes a dense model (1M, 270M, and 1B) with a context length of 64K. 
We designed a new hybrid architecture that combines a convolutional key-value gated mixer with local-global attention to reduce the KV-cache memory that tends to increase exponentially with long context. 
This architecture delivers up to $\times$4.9 faster prefill and decoding speeds on the NPUs compared to models of a similar scale models.
From a 2T tokens candidate corpus, Opt.Gear is trained on a curated 0.5T tokens subset without knowledge distillation.
This is the most data-efficient of the existing foundation models.
All models are released with open weights and deployment binaries for ONNX, Qualcomm NPU, and Apple ANE making Opt.Gear a practical base for edge applications that need fast, memory-efficient inference and strong task capabilities. Furthermore, to expand the ecosystem of on-device generative language models, we are introducing the Opt.Gear-1M that can be deployed on Micro-Controller Units~(MCUs), a Tiny Language Model (TLM). Opt.Gear-1M is the first generative language model to achieve 20 TPS with W4A32 quantization on the ARM Cortex-M7 CPU of the STM32H747I-DISCO.
\end{abstract}
}

\correspondence{\email{contact@opt-ai.kr}}
\checkdata[HuggingFace]{\url{https://huggingface.co/OptGear}}

\begin{document}

\maketitle
\renewcommand{\thefootnote}{}
\footnotetext{$^\dagger$Opt.Gear Team: Juneyoung Park <jyoung.park@opt-ai.kr>, Youngwook Kwon <ywook.kwon@opt-ai.kr>}
\renewcommand{\thefootnote}{\arabic{footnote}}


\vspace{-1.5em}

\begin{figure}[htbp]
    \centering
    \includegraphics[
        width=\linewidth
    ]{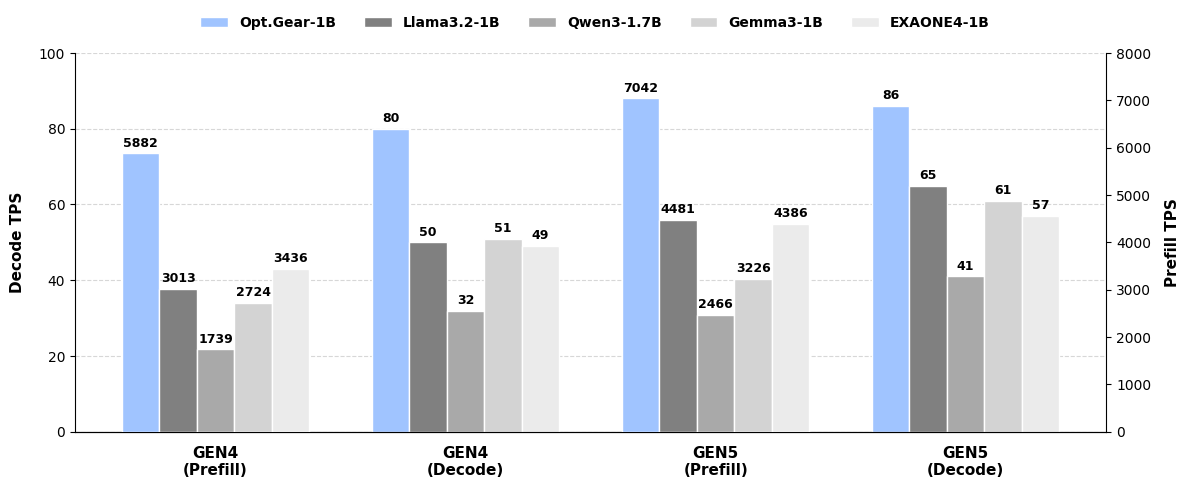}
    \caption{\textbf{Decoder throughput benchmark of Opt.Gear-1B and its counterparts. For hardware evaluation, we report Prefill and Decode TPS across GEN4 and GEN5 using a model compiled with QAIRT.}}
    \label{fig:tps}
\end{figure}

\newpage

\begin{spacing}{0.9}
\tableofcontents
\end{spacing}

\newpage

\section{Introduction}
\label{sec:intro}
On-device LLM applications running directly on smartphones, tablets, laptops, vehicles, robots, and embedded devices can provide low-latency responses, improved privacy, offline availability, and more predictable user experiences. 
However, these advantages come with strict system constraints. 
On-device models must operate under constrained SRAM, power efficiency, and real-time latency requirements across heterogeneous SoC environments that include CPUs, GPUs, and NPUs.

Recent open-source language models~\citep{grattafiori2024llama3herdmodels, gemmateam2025gemma3technicalreport, yang2025qwen3technicalreport, bae2026exaone40unifiedlarge, amini2025lfm2technicalreport} have achieved strong performance by scaling to billions or tens of billions of parameters and by optimizing inference efficiency for server-class GPUs.  
However, these models are designed on the assumption that high-capacity, high-bandwidth memory and high-performance accelerators are available. 
It remains a challenge to deploy these models directly on mobile and edge devices. 
Efficient model architecture design requires not only reducing the number of parameters but also optimized complex trade-offs between hardware runtime profiles, memory hierarchies, kernel compatibility, quantized variants suited to edge runtimes, and the significant computational cost of long-context inference.

The Grouped-Query Attention~(GQA)~\cite{ainslie2023gqatraininggeneralizedmultiquery} is one of the major bottlenecks in on-device LLM inference. 
It has complexity of $O(N^2)$ with sequence length, adopting sliding window attention does not completely resolve this problem. 
The fundamental requirements of computing QK scores, applying attention masks, performing softmax normalization continue to exert heavy pressure on device memory and compute resources.
Furthermore, auto-regressive decoding requires the KV-cache to be stored and accessed repeatedly for each attention layer.
The memory usage and memory bandwidth requirements of the KV-cache increase proportionally with the context length. 
This is a significant bottleneck for latency and power consumption in mobile NPUs and GPUs, which have limited SRAM capacity and memory bandwidth.

We introduce first generation of OptAI Foundation Models.
It is optimized for on-device deployment.
The Opt.Gear is not to build smaller language models, but to jointly improve downstream quality, device-side latency, memory efficiency, long-context capability, and practical deployability under hardware constraints. 
To achieve this, Opt.Gear adapt a hybrid architecture that replaced sliding attenetion to a softmax and matrix multiplication free local mixing module. 
This reduces the overhead of GQA computation and KV-cache access while preserving the representation.

The core of Opt.Gear architecture is the ConvKV-Gated Mixer. 
It uses causal convolution to the key and value, and gates the values based on query-key interactions.
The ConvKV-Gated Mixer does not need a sequence length dependent KV cache and reduces the KV cache capacity during long context decoding while maintaining a small, fixed-size convolution state.
Opt.Gear also adopts GQA that preserves global information routing and long-range context aggregation. 
This hybrid architecture allows Opt.Gear to improve the inference efficiency while maintaining downstream quality.

Opt.Gear was designed with a bilingual model that prioritizes the Korean language.
We adopted the KORMo~\cite{kim2025kormokoreanopenreasoning} tokenizer to improve tokenization efficiency for Korean and English. 
This is followed by Korean-English pretraining, long-context extension, and supervised fine-tuning. 
This multi-stage training process, Opt.Gear aims to bilingual understanding, mathematical and scientific reasoning, long-document comprehension, and instruction-following behavior.

Opt.Gear is optimized for deployment compatibility and supports inference with Transformers~\cite{wolf2020huggingfacestransformersstateoftheartnatural}, llama.cpp~\cite{gerganov_llama_cpp}, ExecuTorch, and vLLM~\cite{kwon2023efficientmemorymanagementlarge}.
Furthermore, executable binary files optimized for NPU inference~(Qualcomm Hexagon NPU, Apple ANE) are also provided.

The key aspects of the Opt.Gear family are summarized as follows:
\begin{itemize}

    \item \textbf{Data-efficient training}: Opt.Gear is trained on a curated 0.5T tokens subset selected from a 2T tokens candidate corpus, without using knowledge distillation from a teacher model.

    \item \textbf{Hardware-aware architecture design}: Opt.Gear is designed to be deployed on diverse device solutions, ranging from MCUs, mobile devices, and GPUs.

    \item \textbf{MCU-scale generation}: Opt.Gear-1M extends the Opt.Gear design to microcontroller-class devices, targeting sub-1 MB deployment with W4A32 quantization on the ARM Cortex-M7 CPU.

    \item \textbf{NPU-friendly optimization}: Opt.Gear replaces excessively used, unnecessary dynamic QKV matrix multiplication and softmax normalization operations with static linear, convolution, and element-wise operations.

    \item \textbf{Reduced KV-cache capacity}: Opt.Gear replaces the sequence-length-dependent KV-cache of local attention layers with a small fixed-size convolutional state.
    
\end{itemize}

In this report, We describe the detail of the Opt.Gear backbone design, training, and deployment optimization.
Finally, we discuss related work, limitations, future directions, and the broader implications of Opt.Gear for edge deployment.

\section{Architecture}
\label{sec:arch}
\begin{figure}[htbp]
    \centering
    \includegraphics[width=1\textwidth,
        trim={0 4cm 0 5cm}, clip]{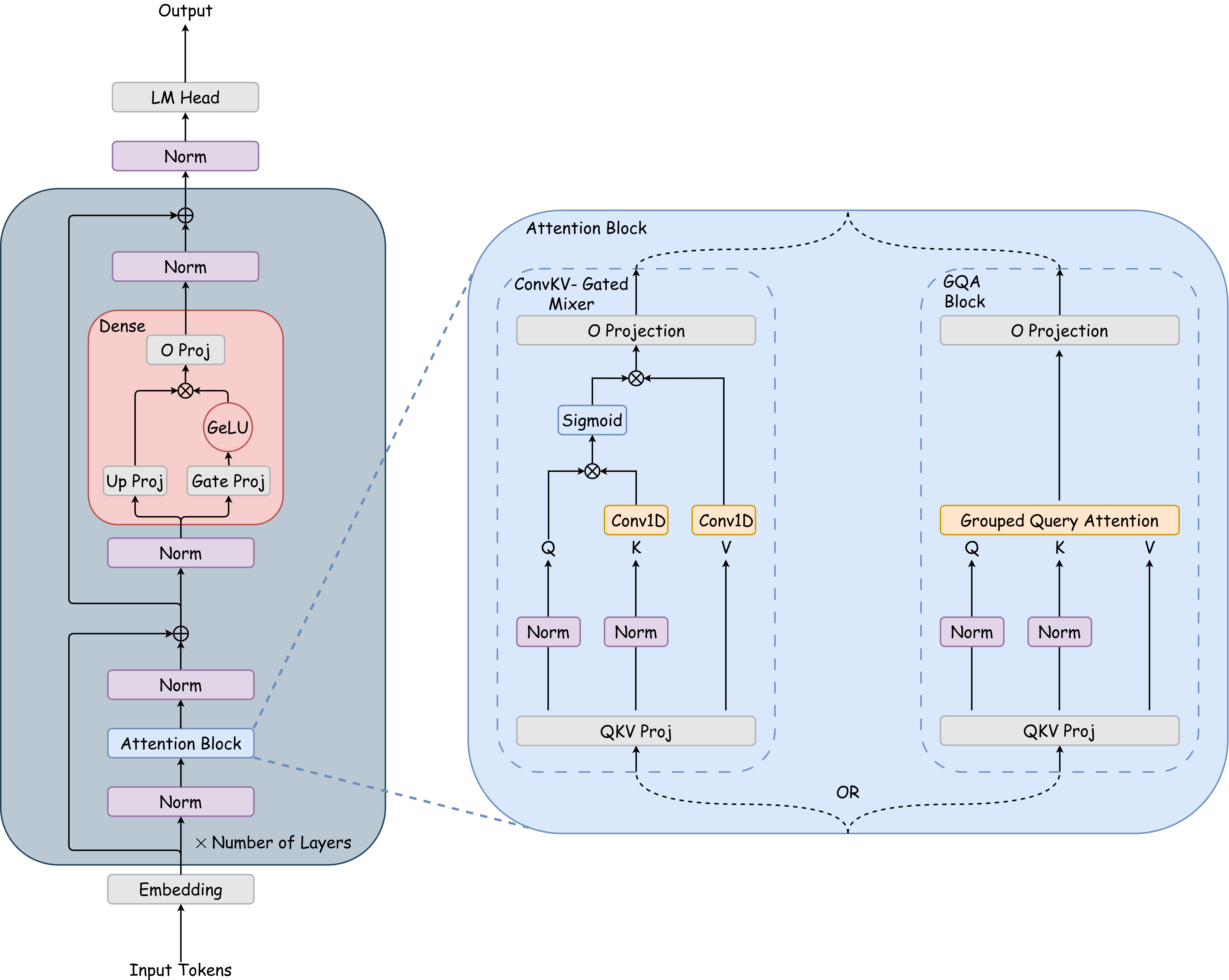}
    \caption{\textbf{Opt.Gear architecture.}}
    \label{fig:Opt.Gear_architecture}
\end{figure}

We introduced in Section~\ref{sec:intro} with hybrid architecture optimized under hardware constraints.
The 3 dense model are released: Opt.Gear-1M, Opt.Gear-270M, and Opt.Gear-1B.
In figure~\ref{fig:Opt.Gear_architecture}, Opt.Gear consists of GQA, ConvKV-Gated Mixer, dense modules.
In this section, we describe the architecture design procedure, and the Opt.Gear-270M, Opt.Gear-1B configurations.

\subsection{Basic Architecture}
\label{subsec:basic_arch}

The core design objective of Opt.Gear is to reduce decoding-time memory traffic while preserving enough global context capacity for long-context tasks. 
Full attention provides strong content-addressable retrieval, but it requires repeated access to key/value states during decoding.
In contrast, recurrent, state-space, and linear-attention style mechanisms~\citep{dao2024transformersssmsgeneralizedmodels, waleffe2024empiricalstudymambabasedlanguage, nvidia2025nemotronhfamilyaccurateefficient, yang2025gateddeltanetworksimproving, kimiteam2025kimilinearexpressiveefficient} keep a bounded state, but they can weaken long-range retrieval and often rely on specialized kernels for their best speedups. 
Opt.Gear therefore uses a hybrid layout: a small set of attention layers handles global routing and mid-range context~\cite{child2019generatinglongsequencessparse}, while ConvKV-Gated Mixer blocks provide the common local mixing path using operations that are widely supported on CPUs, GPUs, and NPUs.

\subsubsection{ConvKV-Gated Mixer}
\label{sec:convkv_gated_mixer}

The ConvKV-Gated Mixer is the primary local sequence operator used in Opt.Gear. For an input hidden sequence \(\mathbf{X} = [x_1, \ldots, x_T] \in \mathbb{R}^{T \times d}\), each token is first projected into query, key, and value streams:
\[
q_t = x_t W_q,\quad k_t = x_t W_k,\quad v_t = x_t W_v.
\]
The key and value streams are then mixed with causal depthwise 1D convolutions:
\[
\widetilde{k}_t = \sum_{i=0}^{L_{\mathrm{conv}}-1} a_i \odot k_{t-i},\quad
\widetilde{v}_t = \sum_{i=0}^{L_{\mathrm{conv}}-1} b_i \odot v_{t-i},
\]
where \(L_{\mathrm{conv}}\) is the convolutional cache length, \(a_i\) and \(b_i\) are channel-wise convolution weights, and invalid past positions are masked by causal padding.

The convolved key stream modulates the current query through an element-wise gate, and the resulting gate selects information from the convolved value stream:
\[
g_t = \sigma(q_t \odot \widetilde{k}_t),\quad
y_t = g_t \odot \widetilde{v}_t,\quad
o_t = y_t W_o.
\]
Here \(\odot\) denotes element-wise multiplication and \(\sigma\) denotes the sigmoid function. This gives the mixer a content-dependent update, similar in spirit to attention, but without constructing a token-by-token attention score matrix or applying softmax over a local window.

The block is intentionally local. It is not meant to replace all attention layers; instead, it replaces many sliding-window local attention layers whose main role is short-range context mixing. 
Long-range retrieval is assigned to the remaining GQA layers, while ConvKV-Gated Mixer layers provide low-cost local interaction through dense projections, causal depthwise convolution, element-wise gating, and an output projection. 
This design makes the layer easier to map onto edge runtimes than operators that require custom sequence kernels.

\subsubsection{Memory-Efficient Caching Mechanism}
\label{sec:memory_efficient_cache}

In on-device inference, system performance is often limited not only by arithmetic throughput, but also by the cost of memory access. 
During auto-regressive decoding, each generated token requires the runtime to read persistent states from previous tokens and write the updated states back to memory. 
When these states no longer fit in on-chip SRAM, external DRAM traffic can become a major source of latency and power consumption.

The persistent state is the KV-cache of GQA. 
Let \(C\) denote the context length, \(W\) the sliding-window size, and \(d_{\mathrm{kv}}\) the key/value dimension. 
A GQA stores key and value vectors for all previous tokens, while a sliding-window attention stores key and value vectors for the most recent \(W\) tokens:
\[
\mathrm{Cache}_{\mathrm{global}} = O(2 C d_{\mathrm{kv}}), \quad
\mathrm{Cache}_{\mathrm{local}} = O(2 W d_{\mathrm{kv}}).
\]
The factor of two accounts for the key and value streams. 
Thus, even when local attention bounds the cache by a fixed window, increasing the window size directly increases the amount of live decoding state.

ConvKV-Gated Mixer changes this local-cache pattern. 
It does not keep a token window for pairwise attention scoring. 
Instead, it keeps a fixed convolution buffer for the projected key and value streams:
\[
\mathcal{C}_K, \mathcal{C}_V \in \mathbb{R}^{d_{\mathrm{kv}} \times L_{\mathrm{conv}}},
\quad
\mathrm{Cache}_{\mathrm{ConvKV}} = O(2 L_{\mathrm{conv}} d_{\mathrm{kv}}).
\]
Here \(L_{\mathrm{conv}}\) is determined by the convolution kernel size. 
This buffer is persistent decoding state, but it is a convolutional state rather than an attention KV-cache. 
After prefill, only the final \(L_{\mathrm{conv}}\) projected key/value states need to remain live for each ConvKV layer. 
During decoding, the runtime appends the current key/value projections, drops the oldest entries, applies the depthwise convolution over the fixed buffer, and computes the gate for the current token.

This mechanism does not eliminate the KV-cache from the entire model: the remaining GQA layers still maintain attention caches for global information routing. 
The saving comes from replacing many local attention layers, whose cache scales with \(W\), with ConvKV-Gated Mixer layers, whose persistent state scales only with \(L_{\mathrm{conv}}\). 
In Opt.Gear-270M and Opt.Gear-1B, \(L_{\mathrm{conv}}=3\); in Opt.Gear-1M, \(L_{\mathrm{conv}}=4\). Because these values are much smaller than the sliding-window and independent of the context length, the mixer substantially reduces the live state and memory bandwidth required for local sequence modeling during long-context decoding.



\subsubsection{Model Configuration}

\begin{table}[htbp]
    \centering
    \caption{Configurations of Opt.Gear.}
    \small
    \setlength{\doublerulesep}{1pt}
    \begin{tabular}{l|ccc}
        \toprule
        Model size & 1B & 270M & 1M \\
        \midrule
        $d$\_model & 1,152 & 640 &  128 \\
        Number of layers & 26 & 18 & 5 \\
        Normalization & QK-LN & QK-LN &  QK-LN \\
        \midrule
        Non-linearity & GeGLU & GeGLU & ReLU6GLU \\
        Feedforward dimension & 6,912 & 2,048 & 272 \\
        \midrule
        Attention type & Hybrid & Hybrid & Hybrid \\
        Head type & GQA & GQA & GQA \\
        Number of heads & 4 & 4 & 2 \\
        Number of KV heads & 1 & 1 & 1 \\
        Sliding-window size & 512 & 512 & - \\
        Head size & 256 & 256 & 64 \\
        Conv kernel size & 3 & 3 & 4 \\
        Max sequence length & 65,536 & 65,536 & 512 \\
        RoPE global theta & 1,000,000 & 1,000,000 & 1,000,000 \\
        RoPE local theta & 10,000 & 10,000 & - \\
        \midrule
        Tokenizer & KORMo & KORMo & BPE \\
        Vocab size & 125,184 & 125,184 & 2,048 \\
        Tied word embedding & False & False & False \\
        \bottomrule
    \end{tabular}
    \label{tab:Opt.Gear_cfg}
\end{table}

The Opt.Gear model employs a hybrid sequence mixing architecture that combines global attention, local attention, and the ConvKV-Gated Mixer. 
The overall configuration of Opt.Gear-1B, Opt.Gear-270M, Opt.Gear-1M is summarized in Table~\ref{tab:Opt.Gear_cfg}. 

Opt.Gear keeps a small number of global attention layers and combines them with local attention and ConvKV-Gated Mixer layers. 
Global attention layers provide long-context context, local attention layers preserve mid-range context within a fixed window, and ConvKV-Gated Mixer layers focus on the short-range context.

Opt.Gear-1B and Opt.Gear-270M use grouped-query attention~(GQA) with four query heads and one key-value head. The head dimension is set to 256, and the sliding-window size is fixed to 512. Opt.Gear-1M uses a smaller GQA configuration with two query heads, one key-value head, and a head dimension of 64. This configuration reduces key-value cache capacity compared with multi-head attention while preserving multiple query heads for expressive token interaction.

The ConvKV-Gated Mixer uses causal depthwise convolution and keeps fixed-size convolutional states for the key and value. Opt.Gear-1B and Opt.Gear-270M use a convolution kernel size of 3, while Opt.Gear-1M uses a kernel size of 4 to increase the local receptive field within a very small hidden dimension. Opt.Gear also adopts query-key normalization~(QK-LN)~\cite{dehghani2023scalingvisiontransformers22}. The feed-forward network follows a dense gated MLP design: Opt.Gear-1B and Opt.Gear-270M use GeGLU~\cite{shazeer2020gluvariantsimprovetransformer}, while Opt.Gear-1M uses ReLU6GLU. Opt.Gear-1B uses a hidden dimension of 1,152, 26 decoder layers, and a feed-forward dimension of 6,912. Opt.Gear-270M uses a hidden dimension of 640, 18 decoder layers, and a feed-forward dimension of 2,048. Opt.Gear-1M uses a hidden dimension of 128, 5 decoder layers, and a feed-forward dimension of 272.

For positional encoding, Opt.Gear uses rotary position embeddings~\cite{su2023roformerenhancedtransformerrotary}. Opt.Gear-1B and Opt.Gear-270M use separate global and local base frequencies: the global RoPE theta is set to 1,000,000, while the local RoPE theta is set to 10,000. This configuration is used together with the hybrid attention layout to support long-context modeling up to 65,536 tokens. 
Opt.Gear-1M keeps the global RoPE theta of 1,000,000 but targets a shorter maximum sequence length of 512 tokens, reflecting the SRAM and latency constraints of MCU deployment.

Opt.Gear-1B and Opt.Gear-270M use the KORMo tokenizer with a vocabulary size of 125,184. The tokenizer is designed to support Korean-English bilingual modeling, and the same vocabulary is shared across Opt.Gear-1B and Opt.Gear-270M. 
Opt.Gear-1M uses a custom byte-level BPE tokenizer~\cite{sennrich2016neuralmachinetranslationrare} with a vocabulary size of 2,048. 
This tokenizer supports compact English-language generation and is designed to keep the embedding tables and runtime memory footprint compatible with flash-limited deployment.
We use untied word embeddings in all configurations, allowing the input embedding table and language modeling head to adapt independently during pretraining and instruction tuning.

Overall, Opt.Gear is designed as a practical dense model family for on-device inference. 
Opt.Gear-1B targets a stronger quality-efficiency trade-off, while Opt.Gear-270M targets lower-latency deployment on more constrained mobile and edge devices. 
Both configurations share the same hybrid design and support 64K context length, enabling long-context applications while reducing the local attention cache footprint. 
Opt.Gear-1M extends the family to MCU-class deployment, where the primary objective is not broad benchmark coverage but stable generative capability under extremely small flash, SRAM, and compute budgets.


Opt.Gear-1M is designed for the lower end of on-device generative modeling, where the target hardware is a microcontroller rather than a mobile NPU or server GPU. 
This setting changes the optimization problem substantially. 
Flash capacity limits the number of parameters and tokenizer size, SRAM limits activation and cache storage, and the absence of large matrix accelerators makes operator simplicity more important than peak arithmetic throughput. 
Opt.Gear-1M therefore keeps the same design philosophy as the larger Opt.Gear models but scales every component around MCU constraints.

The model uses a 128-dim hidden state, 5 decoder layers, 2 GQA query heads, a single key-value head, and a 2,048 token BPE vocabulary. 
The 512 context length is chosen to support short interactive generation and command-style use cases while keeping cache and activation storage bounded. 
The ConvKV-Gated Mixer is especially useful in this regime because its state size is fixed by the convolution kernel rather than by the full context length. 
This avoids maintaining a large sequence-dependent cache on SRAM-limited hardware.

For deployment, Opt.Gear-1M is evaluated with W4A32 quantization on the ARM Cortex-M7 CPU of the STM32H747I-DISCO. 
Under this configuration, Opt.Gear-1M reaches 20 tokens per second, demonstrating that an auto-regressive generative language model can run interactively on an MCU device.
This result expands the deployment target of the Opt.Gear family from mobile and edge accelerators to deeply embedded systems such as sensors, controllers, small robots, and offline human-machine interfaces.

\subsubsection{Device-Side Inference Analysis}

We evaluate Opt.Gear-1M directly on the STM32H747I-DISCO rather than using a host-side simulator. 
The deployment target is the STM32H747XIH6 MCU, and the Cortex-M7 core is configured at 400 MHz. 
The measured binary uses a embedded C runtime generated for STM32CubeIDE and runs without depending on an ONNX~\cite{onnx} graph or the X-CUBE-AI~\cite{x-cube-ai} generated network. 
The runtime uses 4-bit weights, 16-bit embeddings and LM head, and 32-bit floating-point activations, corresponding to the W4A32 deployment setting used in this report.

The device-side timing path is instrumented inside the firmware with \texttt{HAL\_GetTick}. The prefill time is measured around the full prompt forward pass, and decoding time is measured around the per-token decode function. 
The firmware accumulates the number of generated tokens, prefill latency, decode latency, total inference latency, and time-to-first-token~(TTFT) across the built-in prompt set, and then prints a \texttt{timing[total]} line through UART. 
Therefore, the reported throughput is measured from the actual Cortex-M7 execution path and reflects the model-compute portion of auto-regressive generation on the device.

\begin{table}[htbp]
\centering
\caption{\textbf{Opt.Gear-1M device-side inference and build footprint on STM32H747I-DISCO.} The per-token latency is derived from the measured throughput. Loadable section sizes are measured from the final STM32 ELF with \texttt{arm-none-eabi-size}.}
\label{tab:Opt.Gear1m_device_inference}
\small
\setlength{\tabcolsep}{5pt}
\begin{tabular}{ll}
\toprule
\textbf{Item}                        & \textbf{Measured or configured value} \\
\midrule
Board                                & STM32H747I-DISCO \\
MCU / core                           & STM32H747XIH6 / ARM Cortex-M7 \\
Configured CPU clock                 & 400 MHz \\
Quantized format                     & W4A32~(q4 weights, q16 embedding/LM head) \\
Measured throughput                  & 20 tokens/s~(50 ms/token) \\
.text / .rodata section              & 1,596,400 bytes \\
.data section                        & 532 bytes \\
.bss section                         & 444,404 bytes \\
Static RAM heap / stack              & 449,032 bytes \\
\bottomrule
\end{tabular}
\end{table}

Table~\ref{tab:Opt.Gear1m_device_inference} shows that the measured STM32 deployment is limited by both compute throughput and static memory allocation. 
The measured throughput of 20 tokens/s corresponds to roughly 50 ms per generated token, which is fast enough for short command interaction and simple local assistant behavior. 
This is an important threshold for MCU deployment: the model does not merely fit in the build, but also generates at a rate where a user can observe token-by-token output without a server or mobile accelerator.

The build footprint also illustrates the main deployment trade-off. 
For the final ELF, \texttt{arm-none-eabi-size} reports 1,596,400 bytes of .text/.rodata, 532 bytes of .data, and 444,404 bytes of .bss. 
Including the reserved heap/stack region, the static RAM footprint is 449,032 bytes.
This footprint is dominated by statically allocated activation, cache, and intermediate buffers in the embedded C runtime. In particular, the measured deployment build uses a 160-token sequence length and 40-token generation limit, even though the model configuration targets a longer maximum context. 
This choice keeps the auto-regressive cache and workspace bounded on the MCU while preserving enough context for short instructions and embedded-control prompts.

These measurements confirm the practical role of the ConvKV-Gated Mixer in the Opt.Gear-1M. 
On a Cortex-M7 CPU, the expensive part of local sequence modeling is not only arithmetic, but also the movement and retention of state in a very small memory hierarchy. 
By using fixed-size convolutional states for local mixing and keeping only a small GQA cache for attention layers, Opt.Gear-1M reduces the amount of sequence-dependent state that must remain live during decoding. 
The resulting design is suitable for short, local generation on embedded systems, where predictable memory use and stable interactive latency are more important than long-context benchmark performance.







\subsubsection{Quantization-Aware Training}
\label{sec:qat}

Low-bit quantization reduces model size and computational cost, but often causes significant accuracy degradation. 
This degradation is determined not only by the bit-width but also by the quantization range, scale granularity, and discrete quantization grid. 
Post-training quantization cannot compensate for the resulting perturbations because the pretrained weights remain fixed. Quantization-aware training~(QAT)~\cite{jacob2017quantizationtrainingneuralnetworks} addresses this limitation by allowing the model to adapt to quantization during training.

Conventional QAT typically initializes the quantization scale using simple weight statistics, such as the maximum absolute value. 
However, this approach treats all weight perturbations equally, even though their effects on the layer output depend on the input distribution. 
We therefore use calibration activations to estimate the sensitivity of each layer and initialize the group-wise INT4 scales accordingly.

Following GPTQ~\cite{frantar2023gptqaccurateposttrainingquantization}, we formulate quantization as a layer-wise output reconstruction problem. For each quantization group, the initial scale is selected as

\begin{equation}
    \Delta^{*}
    =
    \underset{\Delta}{\arg\min}
    \left\|
        WX-Q_{\Delta}(W)X
    \right\|_{F}^{2},
    \label{eq:hessian_scale}
\end{equation}

where (W) is the full-precision weight matrix, (X) contains the calibration inputs, and ($Q_{\Delta}(W)$) denotes the weights quantized with scale ($\Delta$). 
This objective directly minimizes the discrepancy between the full-precision and quantized layer outputs. The curvature of the reconstruction objective with respect to weight perturbations is governed by the input correlation matrix ($XX^{\top}$). Consequently, the scale initialization accounts for the relative importance of different input directions, rather than treating all weight perturbations equally. 
We perform this initialization independently for symmetric INT4 groups of 32 weights.

The resulting scales are subsequently refined using Learned Step Size Quantization~(LSQ)~\cite{esser2020learnedstepsizequantization}. 
Specifically, the scale ($\Delta_g$) of each quantization group (g) is treated as a learnable parameter, and its weights are fake-quantized during the forward pass as

\begin{equation}
\widehat{W}*{g}
=
\Delta_g
\cdot
\operatorname{clip}
\left(
\operatorname{round}
\left(
\frac{W_g}{\Delta_g}
\right),
q*{\min},
q_{\max}
\right),
\label{eq:lsq_quantizer}
\end{equation}

where ($q_{\min}$) and ($q_{\max}$) define the final symmetric INT4 quantization grid. The non-differentiable rounding operation is approximated using a straight-through estimator during backpropagation. 
Following LSQ, the scale gradients are normalized according to the number of elements in each group and the number of positive quantization levels. 
This normalization prevents the scale updates from dominating the weight updates and enables the group-wise step sizes to be optimized jointly with the model parameters.

Starting from the Hessian-based initialization, QAT jointly refines the full-precision weights and the LSQ step sizes under the global training objective, while the forward pass consistently emulates the final W4A16 deployment format. 
The Hessian-based initialization provides scales with low local output distortion, whereas LSQ allows these scales to adapt to the quantization errors accumulated across layers and to their effect on the final task loss. 
Using the same group-wise INT4 grid during initialization, QAT, and final weight conversion also avoids a quantizer mismatch that could otherwise discard the learned scale structure at deployment time.

\section{Pre-Training}

\subsection{Data Construction}
We first collect a large candidate corpus of approximately 2T tokens from English, Korean, mathematical, and general web sources. Rather than using the entire collected corpus, we apply filtering, deduplication, and mixture rebalancing to construct a curated 0.5T tokens training set. 
This design prioritizes data quality and Korean-English balance over raw token count, allowing Opt.Gear to achieve competitive performance under a limited compute budget without teacher-model distillation.
The Opt.Gear models are pre-trained on a mixture comprising roughly 92\% English text, 6\% Korean text, and 2\% math. 
Notably, we did not train the model on code data, specifically considering that models of the 270M and 1B are not used for code generation tasks.

The KORMo tokenizer employs a byte-level BPE (Byte Pair Encoding) algorithm. Compared to other commercial tokenizers, it maintains a similar performance in English while significantly outperforming commercial models in Korean.

Opt.Gear-1M follows a separate tokenizer and data-design objective. 
Since the model targets MCU deployment, it uses a much smaller 2,048 tokens BPE vocabulary and prioritizes compact English generation, short-form command following, and predictable structured outputs. 
Unlike Opt.Gear-270M, Opt.Gear-1B, and Opt.Gear-1M is not trained with the general pre-training corpus in this stage. 
We do not conduct a separate web-scale pre-training phase for Opt.Gear-1M, its task-specific learning procedure is described in the post-training section.

\subsection{Training Hyper-Parameters}
We employ the AdamW~\cite{adamW} optimizer with hyper-parameters set to $\beta_1=0.9$, $\beta_2=0.95$, and $\mathrm{weight\_decay}=0.1$. 
We set the maximum sequence length to 4K during pre-training, and pre-train Opt.Gear on 0.5T tokens. 
As for the learning rate scheduling, we first linearly increase it from 0 to $1.0 \times 10^{-4}$ during the first 2K steps. 
Then, we keep a constant learning rate of $1 \times 10^{-4}$ until the model consumes 2T training tokens. 
Subsequently, we also linearly decay the learning rate to $1 \times 10^{-5}$ during the last 2K steps. 
The gradient clipping norm is set to 1.0.

\subsection{Long Context Extension}
We adopt a similar approach to KORMo to enable long context capabilities in Opt.Gear. 
After the pre-training stage, we apply prolong-data-64k~\cite{gao2025trainlongcontextlanguagemodels} corpus for context extension and perform two additional training phases, each comprising 10B tokens in both English and Korean, to progressively expand the context window from 4K to 32K and then to 64K. 

\subsection{Infrastructure}
Each node in the H200~\cite{h200} cluster contains 8 GPUs connected by NVLink and NVSwitch within nodes. 
The training of Opt.Gear is supported by the Nanotron~\cite{nanotron} framework, supporting 3D parallelism, and performance optimizations for large-scale training workloads.
On the whole, Opt.Gear applies 8-way Pipeline Parallelism (PP), 1-way Tensor Parallelism (TP), and ZeRO-1 Data Parallelism (DP).

\section{Post-Training}

\subsection{Supervised Fine-Tuning}
Post-training adapts the base Opt.Gear models to instruction-following behavior while keeping the deployment constraints of small on-device models in mind. 
For Opt.Gear-270M and Opt.Gear-1B, we first apply general Supervised Fine-Tuning~(SFT) to improve helpfulness, formatting reliability, and dialogue-style instruction following. 
We then apply reasoning-oriented SFT to improve step-by-step mathematical and analytical responses. 
This two-stage design separates broad instruction alignment from reasoning-format specialization, reducing the risk that reasoning-style data overwhelms everyday assistant behavior.

Opt.Gear Base and Instruction models share the same tokenizer, with additional control tokens dedicated to instruction formatting. 
A key difference is that Base models output an \texttt{<EOS>} token at the end of generation, while Instruction models output \texttt{<EOT>} at the end of the generation. 
Fine-tuning either model type therefore requires adding the corresponding end token and ensuring that the training examples terminate with the same convention used by the target runtime. 
This is particularly important for on-device inference, where malformed termination can waste decoding steps and increase latency.

During SFT data construction, we prioritize concise answers, deterministic formatting, and task coverage that is useful in local assistant scenarios. 
The post-training mixture includes general instruction following, Korean-English bilingual dialogue, summarization, rewriting, mathematical reasoning, and safety-aware refusal examples. 
We avoid relying on teacher-model distillation as a required training signal; instead, post-training is used to shape output format, task controllability, and response style on top of the pre-trained Opt.Gear-270M and Opt.Gear-1B models.

\subsubsection{Opt.Gear-1M Board-Control Post-Training}

Opt.Gear-1M follows a different post-training path from Opt.Gear-270M and Opt.Gear-1B.
Instead, Opt.Gear-1M is trained for a narrow embedded-control setting in which the model maps short natural-language commands to structured board-control intents. 
This design intentionally favors predictable command generation over broad open-domain language ability, matching the constraints of MCU deployment.

To specialize Opt.Gear-1M for STM32 board-control use cases, we construct a GPT-generated synthetic command dataset, denoted as \texttt{stm32\_cmd\_synth}. 
The dataset contains 50,000 prompt-response pairs. 
Each prompt is a short natural-language command for controlling the STM32 board, and each target response is a compact JSON-style control intent. 
The action space covers LCD, LED, and camera control. 
We use this dataset in the Opt.Gear-1M post-training stage so that the model learns to translate informal user requests into structured device-control commands rather than producing unconstrained free-form text.

\subsection{Evaluations}

\subsubsection{Evaluation Benchmarks}
Small models often fail evaluations because they struggle to output answers in the expected format.
Therefore, to ensure a fair evaluation, we conducted our assessments using the LM Evaluation Harness~\cite{eval-harness}.

The base model of Opt.Gear is pretrained on a bilingual corpus with English and Korean constituting the majority, so we evaluate its performance on a series of benchmarks primarily in English and Korean.
Considered benchmarks are categorized and listed as follows:

\textbf{Multi-subject multiple-choice} datasets include MMLU~\cite{mmlu}, MMLU-Pro~\cite{mmlu_pro}, KMMLU~\cite{kmmlu}, and GPQA~\cite{gpqa}.

\textbf{Language understanding and reasoning} datasets include WinoGrande~\cite{winogrande}, HellaSwag~\cite{hellaswag}, PIQA~\cite{piqa}, and ARC~\cite{arc}.

\textbf{Korea understanding and culture} datasets include CLIcK~\cite{click}, HAERAE~\cite{haerae}, and KoBEST~\cite{kobest}.

We adopt perplexity-based evaluation for datasets including HellaSwag, PIQA, WinoGrande, MMLU, MMLU-Pro, GPQA, ARC-Easy, ARC-Challenge, KMMLU, and KoBEST. 

\begin{itemize}
    \item Knowledge Capabilities: Opt.Gear demonstrates competitive performance in MMLU, MMLU-Pro, and GPQA relative to baselines matched in size.
    Additionally, as shown in the results, Opt.Gear-1B scores 43.2\% on MMLU and 30.3\% on GPQA, explicitly outperforming both Gemma3-1B and EXAONE 4.0-1.2B.

    \item Language Understanding and Reasoning: Opt.Gear models exhibit robust commonsense and reasoning capabilities. 
    In the 1B–2B parameter range, Opt.Gear-1B achieves 70.8\% on ARC-Easy and 71.2\% on PIQA, outperforming EXAONE 4.0-1.2B. In the sub-1B category, Opt.Gear-270M demonstrates highly competitive reasoning, scoring 67.3\% on PIQA and 59.1\% on ARC-Easy, which successfully surpasses the distilled Gemma3-270M.

    \item Korea Understanding and Culture: Given its bilingual pretraining corpus, Opt.Gear demonstrates distinct superiority in Korean-specific benchmarks. 
    Opt.Gear-1B comprehensively outperforms Gemma3-1B, Llama3.2-1B, and EXAONE 4.0-1.2B across all Korean tasks (KMMLU, KoBEST, CLIcK, and HAERAE). Notably, on HAERAE, Opt.Gear-1B scores 44.2\%, leading Gemma3-1B and Llama3.2-1B by significant margins. 
    This trend is consistent at the tiny model scale, where Opt.Gear-270M records 30.0\% on KMMLU and 51.9\% on KoBEST, beating both SmolLM2-135M and Gemma3-270M.

    \item Training and Data Efficiency: A crucial takeaway from both tables is the exceptional data efficiency of the Opt.Gear models. 
    While baseline models like Gemma3, Llama3.2, LFM2.5, and Qwen3 are trained on massive datasets (ranging from 2T up to 36T tokens) and frequently leverage knowledge distillation, both Opt.Gear-270M and Opt.Gear-1B achieve highly competitive—and often superior—performance utilizing only 0.5T tokens without any distillation.
\end{itemize}

\clearpage
\begin{table}[tb]
\centering
\caption{Performance of tiny language models (135M--1B parameters) across English, and Korean benchmarks. All results are obtained using Language Model Evaluation Harness, and may differ from other reported scores.}
\label{tab:pretrain_models_u1b}
\small
\setlength{\tabcolsep}{4pt}
\begin{tabular}{lccccc}
\toprule
\textbf{Benchmark} & \textbf{Opt.Gear-270M} & \textbf{SmolLM2-135M} & \textbf{Gemma3-270M} & \textbf{LFM2.5-350M} & \textbf{Qwen3-0.6B} \\
\midrule
\# Trained Tokens & 0.5T & 2T & 6T & 28T & 36T \\
Distilled & \xmark & \xmark & \cmark & \cmark & \cmark \\
\midrule
\multicolumn{5}{c}{\textit{English}} \\
\midrule
MMLU             & 25.9 & 25.3 & 26.5 & 41.0 & 47.3 \\
GPQA             & 23.7 & 21.2 & 22.7 & 26.8 & 29.3 \\
ARC-Easy         & 59.1 & 63.3 & 56.1 & 69.0 & 71.4 \\
ARC-Challenge    & 29.5 & 28.8 & 28.0 & 37.1 & 40.2 \\
HellaSwag        & 38.4 & 43.4 & 38.9 & 42.2 & 46.7 \\
PIQA             & 67.3 & 66.5 & 66.9 & 65.8 & 67.4 \\
WinoGrande       & 51.5 & 54.8 & 51.9 & 51.6 & 55.4 \\
OBQA             & 31.6 & 30.8 & 31.6 & 30.4 & 32.4 \\
\midrule
\multicolumn{5}{c}{\textit{Korean}} \\
\midrule
KMMLU            & 30.0 & 29.1 & 28.0 & 31.4 & 35.0 \\
KoBEST           & 51.9 & 48.7 & 50.0 & 51.1 & 54.4 \\
CLIcK            & 27.2 & 22.5 & 27.3 & 24.0 & 36.7 \\
HAERAE           & 21.5 & 18.9 & 20.4 & 24.8 & 37.0 \\
\bottomrule
\end{tabular}
\end{table}

\begin{table}[tb]
\centering
\caption{Performance of small language models (1B--2B parameters) across English, and Korean benchmarks. All results are obtained using Language Model Evaluation Harness, and may differ from other reported scores.}
\label{tab:pretrain_models_u2b}
\small
\setlength{\tabcolsep}{4pt}
\begin{tabular}{lccccccccccc}
\toprule
\textbf{Benchmark} & \textbf{Opt.Gear-1B} & \textbf{Gemma3-1B} & \textbf{Llama3.2-1B} & \textbf{EXAONE 4.0-1.2B} & \textbf{LFM2.5-1.2B} &\textbf{Qwen3-1.7B} \\
\midrule
\# Trained Tokens & 0.5T & 2T & 9T & 12T & 28T & 36T \\
Distilled & \xmark & \cmark & \cmark & \cmark & \cmark & \cmark \\
\midrule
\multicolumn{6}{c}{\textit{English}} \\
\midrule
MMLU             & 43.2 & 39.8 & 46.1 & 37.2 & 51.1 & 60.3 \\
MMLU-Pro         & 11.3 & 13.7 & 19.4 & 42.8 & 19.7 & 43.0 \\
GPQA             & 30.3 & 23.2 & 28.8 & 25.3 & 27.3 & 24.8 \\
ARC-Easy         & 70.8 & 70.0 & 72.4 & 58.3 & 79.2 & 80.0 \\
ARC-Challenge    & 38.6 & 40.1 & 41.7 & 34.0 & 51.0 & 53.0 \\
HellaSwag        & 56.0 & 60.0 & 61.1 & 42.4 & 62.1 & 60.3 \\
PIQA             & 71.2 & 72.5 & 74.8 & 55.4 & 72.9 & 72.4 \\
WinoGrande       & 58.2 & 58.7 & 62.1 & 53.9 & 60.0 & 60.9 \\
OBQA             & 36.8 & 36.4 & 36.2 & 27.8 & 40.8 & 39.4 \\
\midrule
\multicolumn{6}{c}{\textit{Korean}} \\
\midrule
KMMLU            & 36.0 & 30.7 & 29.9 & 32.6 & 29.2 & 41.6 \\
KoBEST           & 60.3 & 59.5 & 51.8 & 50.7 & 59.1 & 62.8 \\
CLIcK            & 39.7 & 37.8 & 30.6 & 32.9 & 38.3 & 49.4 \\
HAERAE           & 44.2 & 35.3 & 32.6 & 30.2 & 33.9 & 52.0 \\
\bottomrule
\end{tabular}
\end{table}
\clearpage

\begin{table}[t]
\centering
\caption{Performance comparison of quantization methods. The metrics are grouped by language (English and Korean) and rounded to one decimal place. The best post-training quantization result for each metric is highlighted in bold.}
\label{tab:qat_benchmark_rounded}
\small 
\renewcommand{\arraystretch}{1.15}
\setlength{\tabcolsep}{6pt} 
\begin{tabular}{@{} l l ccc ccc @{}} 
\toprule
\multirow{2}{*}{\textbf{Model}} & \multirow{2}{*}{\textbf{Method}} & \multicolumn{3}{c}{\textbf{English}} & \multicolumn{3}{c}{\textbf{Korean}} \\
\cmidrule(lr){3-5} \cmidrule(l){6-8} 
& & \textbf{MMLU} & \textbf{HellaSwag} & \textbf{WinoGrande} & \textbf{KMMLU} & \textbf{KoBEST} & \textbf{HAERAE} \\
\midrule

\rowcolor{black!5} \cellcolor{white} \multirow{4}{*}{\textbf{Opt.Gear-1B}}
 & QAT  & 41.3 & 54.0 & 58.3 & 35.0 & 60.4 & 42.7 \\
 & RTN  & 39.8 & 53.9 & 56.9 & 33.8 & 59.8 & 39.9 \\
 & AWQ  & 39.5 & 52.6 & 56.5 & 32.7 & 59.6 & 38.2 \\
 & GPTQ & 39.6 & 53.5 & 57.1 & 34.4 & 59.6 & 41.0 \\
\midrule

\rowcolor{black!5} \cellcolor{white} \multirow{4}{*}{\textbf{Gemma3-1B}}
 & QAT  & 35.8 & 51.4 & 55.0 & 26.9 & 56.2 & 31.2 \\
 & RTN  & 32.5 & 48.5 & 53.8 & 21.2 & 55.1 & 23.6 \\
 & AWQ  & 34.3 & 50.5 & 53.7 & 23.3 & 56.4 & 29.7 \\
 & GPTQ & 34.9 & 49.6 & 54.9 & 23.3 & 55.1 & 27.1 \\
\midrule

\multirow{3}{*}{\textbf{Llama3.2-1B}} 
 & RTN  & 41.1 & 58.6 & 55.7 & 29.2 & 23.9 & 50.6 \\
 & AWQ  & 43.4 & 60.8 & 58.8 & 28.7 & 30.9 & 50.7 \\
 & GPTQ & 43.6 & 61.1 & 57.4 & 29.2 & 28.6 & 50.7 \\
\bottomrule

\multirow{3}{*}{\textbf{EXAONE 4.0-1.2B}} 
 & RTN  & 34.5 & 40.4 & 53.9 & 30.4 & 29.2 & 50.8 \\
 & AWQ  & 24.4 & 25.6 & 51.1 & 19.1 & 46.5 & 20.7 \\
 & GPTQ & 35.3 & 41.6 & 53.9 & 31.5 & 49.9 & 29.5 \\
\bottomrule

\multirow{3}{*}{\textbf{Qwen3-1.7B}} 
 & RTN  & 56.1 & 59.8 & 59.3 & 40.3 & 47.4 & 60.3 \\
 & AWQ  & 57.9 & 59.8 & 59.2 & 40.8 & 46.9 & 61.9 \\
 & GPTQ & 56.2 & 60.1 & 57.9 & 39.5 & 46.6 & 59.2 \\
\bottomrule

\end{tabular}
\end{table}

We also evaluated various quantization methods, applying PTQ techniques (RTN, AWQ~\cite{awq}, GPTQ) directly to the QAT baselines of Opt.Gear-1B and Gemma3-1B. 
Opt.Gear-1B demonstrates remarkable stability within this pipeline. 
On the English benchmark, it experiences only a marginal performance drop, dipping slightly from 41.3\% to 39.8\% after applying RTN. 
Gemma3-1B, conversely, suffers a much more severe degradation, falling from 35.8\% to 32.5\% under the identical quantization process.

Llama3.2-1B, Qwen3-1.7B achieve competitive English PTQ scores, it struggles significantly on Korean tasks. By leveraging its robust QAT foundation prior to PTQ application, Opt.Gear-1B reliably sustains high performance across both languages without experiencing catastrophic forgetting or significant precision-loss degradation.

\subsubsection{Inference Performance}

\paragraph{Experimental Setup.} 
We evaluated practical inference efficiency in on-device environments by conducting a cross-benchmark that included various hardware processors~(CPU, GPU, NPU) and runtimes~(Qualcomm AI Engine Direct~\cite{qairt}, llama.cpp, CoreML~\cite{coreml}). 
As evaluation metrics, we separately measured the speed of Prefill~(Prompt Processing, tokens/sec), which analyzes large amounts of context, and Decode~(Token Generation, tokens/sec), which generates text sequentially. We applied W4A16 quantization in all runtime environments, and configured the embedding and LM head layers to maintain 16-bit precision.

\paragraph{Qualcomm AI Engine Direct~(QAIRT).} 
To achieve the acceleration performance of the Hexagon NPU on the QAIRT pipeline, this experiment performed comprehensive optimizations at the input/output~(I/O), operation, and graph levels.

Opt.Gear-1B demonstrated overwhelming inference performance on Qualcomm Snapdragon chipset. 
On the Snapdragon GEN5, Opt.Gear-1B achieved a prefill speed of 7,042 tokens/sec, significantly outperforming comparable models such as Llama 3.2-1B (4,481 tokens/sec) and Gemma 3-1B (3,226 tokens/sec).

This extreme performance gap suggests that Opt.Gear-1B architecture is highly optimized for efficient utilization of HTP VTCM~(8 MB) in matrix units.
We confirmed that server-grade prompt throughput can be achieved even in mobile environments through bottleneck-free, constrained HTP VTCM access.

\begin{table}[h]
\centering
\caption{On-device NPU inference performance (Prefill / Decode tokens per second) on Qualcomm AI Runtime.}
\label{tab:qualcomm_runtime}
\small
\renewcommand{\arraystretch}{1.15}
\setlength{\tabcolsep}{8pt}
\begin{tabular}{@{} l cc @{}}
\toprule
\multirow{2}{*}{\textbf{Model}} & \multicolumn{2}{c}{\textbf{NPU}} \\
\cmidrule(l){2-3}
& \textbf{Snapdragon GEN5} & \textbf{Snapdragon GEN4} \\
\midrule

\rowcolor{black!5} \cellcolor{white} \textbf{Opt.Gear-1B} & \textbf{7042} / \textbf{86} & \textbf{5882} / \textbf{80} \\
Llama3.2-1B & 4481 / 65 & 3013 / 50 \\
Gemma3-1B   & 3226 / 61 & 2724 / 51 \\
Qwen 3-1.7B & 2466 / 41 & 1739 / 32 \\
EXAONE 4.0-1.2B  & 4386 / 57 & 3436 / 49 \\
\bottomrule
\end{tabular}
\end{table}

\paragraph{llama.cpp.} 
Experimental results in the llama.cpp environment, a general-purpose open-source inference framework, clearly demonstrate the importance of runtime optimization. 
While Opt.Gear-1B maintained excellent and stable performance in CPU and GPU environments, an exceptional sharp drop in performance was observed in the NPU environment (Prefill 290 tokens/sec based on Snapdragon GEN5). 
This appears to be due to a bottleneck occurring because the llama.cpp NPU backend failed to allocate and optimize the Sliding Window Attention pattern using hardware accelerators. 
On the other hand, the LFM 2.5-1.2B model, which is not affected by the structure, recorded 2691 tokens/sec in the same environment and occupied the NPU normally.

\begin{table}[h]
\centering
\caption{Cross-processor inference performance (Prefill / Decode). \textbf{Top:} Qualcomm Snapdragon chipsets using llama.cpp. \textbf{Bottom:} Apple iPhone models using CoreML Runtime. Unavailable metrics are denoted with hyphens.}
\label{tab:llamacpp_coreml_ondevice_performance}
\small
\renewcommand{\arraystretch}{1.15}
\setlength{\tabcolsep}{4.5pt}
\begin{tabular}{@{} l ccc ccc @{}}
\toprule

\multirow{2}{*}{\textbf{Model}} & \multicolumn{3}{c}{\textbf{Snapdragon GEN5}} & \multicolumn{3}{c}{\textbf{Snapdragon GEN4}} \\
\cmidrule(lr){2-4} \cmidrule(l){5-7}
& \textbf{CPU} & \textbf{GPU} & \textbf{NPU} & \textbf{CPU} & \textbf{GPU} & \textbf{NPU} \\
\midrule
\rowcolor{black!5} \cellcolor{white} \textbf{Opt.Gear-1B} & \textbf{686} / 36 & \textbf{1230} / \textbf{64} & 290 / \textbf{65} & \textbf{626} / 30 & \textbf{1118} / \textbf{64} & 217 / \textbf{59} \\
Llama3.2-1B & 470 / 33 & 1061 / 39 & 1912 / 47 & 409 / 29 & 947 / 34 & 1339 / 47 \\
Gemma3-1B & 286 / \textbf{57} & 1201 / 54 & 285 / 54 & 580 / 28 & 1059 / 52 & 215 / 43 \\
LFM2.5-1.2B & 565 / 40 & 1164 / 55 & \textbf{2691} / 61 & 515 / \textbf{33} & 1043 / 51 & \textbf{1972} / 57 \\
EXAONE 4.0-1.2B & 367 / 27 & 796 / 29 & 1315 / 37 & 319 / 20 & 713 / 24 & 895 / 35 \\

\midrule
\addlinespace 

\multirow{2}{*}{\textbf{Model}} & \multicolumn{3}{c}{\textbf{iPhone 17 Pro}} & \multicolumn{3}{c}{\textbf{iPhone 16 Pro}} \\
\cmidrule(lr){2-4} \cmidrule(l){5-7}
& \textbf{CPU} & \textbf{GPU} & \textbf{NPU} & \textbf{CPU} & \textbf{GPU} & \textbf{NPU} \\
\midrule
\rowcolor{black!5} \cellcolor{white} \textbf{Opt.Gear-1B} & \textbf{613} / \textbf{58} & \textbf{2050} / \textbf{98} & \textbf{3085} / \textbf{97} & \textbf{642} / \textbf{56} & 999 / 40 & \textbf{2589} / \textbf{80} \\
Llama3.2-1B & 362 / 41 & 1308 / 74 & 1073 / 34 & 381 / 40 & 615 / 42 & 696 / 25 \\
Gemma3-1B & 544 / 45 & 1852 / 60 & 2380 / 90 & 587 / 46 & \textbf{1010} / 34 & 1862 / 70 \\
LFM2.5-1.2B & 466 / 50 & 1503 / 80 & - / - & 455 / 42 & 679 / \textbf{47} & - / - \\
EXAONE 4.0-1.2B & 286 / 40 & 1192 / 66 & 590 / 20 & 254 / 33 & 379 / 34 & 319 / 15 \\

\bottomrule
\end{tabular}
\end{table}

\paragraph{CoreML Runtime.}
Opt.Gear-1B also demonstrated outstanding versatility by achieving the highest performance in the CoreML. 
In particular, on the iPhone 17 Pro, it delivered significantly higher prefill throughput than other models on both the GPU~(2,050 tokens/sec) and the NPU~(3,085 tokens/sec). 
These results are even more remarkable given the hardware limitations inherent to Apple Silicon. 
The ANE (Apple Neural Engine) is significantly impacted by memory bottlenecks, which have a major negative effect on inference speeds. 
As a result, the Llama 3.2 and EXAONE 4.0 models, which adopt a full-attention architecture, showed significant latency and performance degradation. 
The LFM 2.5-1.2B model was excluded from NPU measurement because there was no CoreML binary.

\subsection{Discussion}

\subsubsection{Training with a Minimal Corpus}

Many recent small language models rely on distillation from larger teacher models to obtain strong benchmark performance with limited parameter counts. 
Distillation can improve sample efficiency, reasoning style, and answer formatting, but it also introduces a large additional training and storage cost. 
Opt.Gear is trained without knowledge distillation~(KD)~\cite{knowledge_distillation} from a teacher model. 
This makes the training recipe simpler to reproduce and keeps the model behavior more directly tied to the curated pre-training and post-training data.

This decision is motivated not only by model-design clarity, but also by practical resource constraints. 
Opt.Gear is trained with a limited setup of 8 $\times$ H200 140 GB GPUs. 
Under this budget, we estimate that even a non-distilled 1T-token training run would require approximately 30 days. 
Online knowledge distillation is substantially more expensive because every student update also requires teacher-model inference. 
We further estimate that a 100B tokens online KD run would already require approximately 30 days, making it impractical to scale to the 0.5T tokens Opt.Gear recipe.

Offline KD is also not practical under this setting. 
Storing teacher logits avoids repeated teacher inference, but the storage requirement becomes the bottleneck. 
When storing teacher Top-\(K=32\) logits, 1 TB of storage can hold only about 6B tokens. 
This is less than 2\% of a 0.5T tokens corpus, and therefore cannot cover the training data at the scale required for Opt.Gear without either repeatedly regenerating logits or maintaining prohibitively large storage.

\begin{table}[htbp]
\centering
\caption{\textbf{Practical constraints of teacher distillation for Opt.Gear.} Estimates are based on the available 8-H200-140GB training setup.}
\label{tab:kd_resource_constraints}
\small
\setlength{\tabcolsep}{4pt}
\begin{tabular}{p{0.27\textwidth}p{0.30\textwidth}p{0.33\textwidth}}
\toprule
\textbf{Training strategy} & \textbf{Estimated requirement} & \textbf{Implication} \\
\midrule
Non-distilled training & 1T tokens require about 30 days & Already compute-intensive under the available hardware budget \\
Online KD & 100B tokens require about 30 days & Too expensive to scale to the 0.5T tokens Opt.Gear recipe \\
Offline KD with Top-\(K=32\) logits & 1 TB stores only about 6B tokens & Storage cannot cover the required training corpus scale \\
\bottomrule
\end{tabular}
\end{table}

This choice creates a different trade-off. 
Without a teacher model, Opt.Gear must acquire knowledge, bilingual capability, and reasoning patterns from the training corpus itself. 
As a result, the quality of data filtering, tokenizer design, and curriculum scheduling becomes especially important. 
The advantage is that the final model provides a clearer measurement of architecture and data efficiency, particularly for Korean-English bilingual modeling and long-context on-device deployment.

\section{Conclusion, Limitations, and Future Directions}
\label{sec:conclusion}

Opt.Gear-270M and Opt.Gear-1B are trained as Korean-English bilingual models, while Opt.Gear-1M extends the same hardware-aware design philosophy to microcontroller-scale English generation. 
For the larger models, the training pipeline combines a limited budget of 0.5T tokens, long-context extension, and instruction tuning without relying on teacher-model distillation. 
For Opt.Gear-1M, the pipeline skips general pre-training and instead uses task-focused training on STM32 board-control commands, together with a compact tokenizer, a 512-token context window, W4A32 deployment, and interactive generation on MCU hardware. This design emphasizes data efficiency, deployability, and transparent measurement of the capabilities provided by the model architecture itself.

Nevertheless, the constrained 0.5T tokens training budget also limits the capabilities of Opt.Gear-270M and Opt.Gear-1B, particularly in complex reasoning and mathematical reasoning tasks that typically benefit from larger and more carefully structured training corpus. 
The current results therefore reflect not only the capacity of the proposed architecture but also the limitations of the available data and compute budget. 
Because data scale, data composition, and model capacity were not independently scaled in this study, the results do not yet establish whether the remaining reasoning gap originates primarily from model size, insufficient task-relevant data, or training inefficiency.

Future work should therefore focus not only on increasing the number of training tokens, but also on improving the capability gained per token. 
Promising directions include higher-quality data selection, domain-aware mixture optimization, curriculum scheduling, targeted reasoning and mathematical data, and training objectives that make more effective use of limited compute. 
Investigating these methods under a fixed data or compute budget will be important for determining how much the reasoning capability of compact models can be improved without sacrificing the deployment efficiency that Opt.Gear is designed to provide.




%

\bibliographystyle{plainnat}
\bibliography{main}
\begingroup
\renewcommand{\thefootnote}{}
\footnotetext{We would like to thank Seongbae Lee, Seongjae Park, and Jinwoo Lee of the OptAI Model Optimization Team for helping bring this research to a successful completion. We also remember the Marcus Building as our own garage, the place where this journey began.}
\endgroup




%
%






\end{document}

%% file: commands.tex
\renewcommand{\phi}{\varphi}

\renewcommand{\epsilon}{\varepsilon}
\renewcommand{\imath}{\mathrm{i}}

\newlength{\restsubwidth}
\newlength{\restsubheight}
\newlength{\restsubmoreheight}
\newcommand{\rest}[2]{%
        \settowidth{\restsubwidth}{\ensuremath{#2}}
        \settoheight{\restsubheight}{\ensuremath{{}_{#2}}}
        \ensuremath{{#1\hskip 0.5pt}_{\vrule\kern2pt\parbox[b][%
        4pt][b]{\the\restsubwidth}{%
                        \ensuremath{{}_{#2}}}}}
        }